\documentclass[a4paper, 10pt, conference]{ieeeconf}      

\IEEEoverridecommandlockouts                              

\usepackage{graphics} 
\usepackage{epsfig} 
\usepackage{mathptmx} 
\usepackage{times} 
\usepackage{amsmath} 
\usepackage{amssymb}  
\usepackage{float}

\title{\LARGE \bf
Scalable, Training-Free Visual Language Robotics: a modular multi-model framework for consumer-grade GPUs
}

\author{Marie Samson$^{*}$, Bastien Muraccioli$^{*}$, Fumio Kanehiro$^{*}$
\thanks{$^{*}$M. Samson, B. Muraccioli share the first authorship. M. Samson, B. Muraccioli, F. Kanehiro are with the CNRS-AIST JRL (Joint Robotics Laboratory), National Institute of Advanced Industrial Science and Technology (AIST), Japan.}%
}

\begin{document}

\maketitle
\thispagestyle{empty}
\pagestyle{empty}

\begin{abstract}

The integration of language instructions with robotic control, particularly through Vision Language Action (VLA) models, has shown significant potential. However, these systems are often hindered by high computational costs, the need for extensive retraining, and limited scalability, making them less accessible for widespread use.

In this paper, we introduce SVLR (Scalable Visual Language Robotics)\footnote{https://scalable-visual-language-robotics.github.io/}, an open-source, modular framework that operates without the need for retraining, providing a scalable solution for robotic control. 
SVLR leverages a combination of lightweight, open-source AI models including the Vision-Language Model (VLM) \textit{Mini-InternVL}, zero-shot image segmentation model \textit{CLIPSeg}, Large Language Model \textit{Phi-3}, and sentence similarity model \textit{all-MiniLM} to process visual and language inputs. These models work together to identify objects in an unknown environment, use them as parameters for task execution, and generate a sequence of actions in response to natural language instructions. A key strength of SVLR is its scalability. The framework allows for easy integration of new robotic tasks and robots by simply adding text descriptions and task definitions, without the need for retraining. This modularity ensures that SVLR can continuously adapt to the latest advancements in AI technologies and support a wide range of robots and tasks.

SVLR operates effectively on an NVIDIA RTX 2070 (mobile) GPU, demonstrating promising performance in executing pick-and-place tasks. While these initial results are encouraging, further evaluation across a broader set of tasks and comparisons with existing VLA models are needed to assess SVLR's generalization capabilities and performance in more complex scenarios.

\end{abstract}

\section{INTRODUCTION}

Vision Language Action (VLA) models, which aim to generate robot actions based on visual information and language instructions, have made significant progress in recent years. Models, such as RT-2-X \cite{rt2x} have demonstrated impressive capabilities in robotic manipulation across a diverse set of tasks. The ability to combine vision, language, and action enables robots to understand and execute tasks in dynamic environments, enhancing their autonomy. This integration is particularly beneficial for non-robotics developers, as it simplifies robot operation through natural language, making robots more accessible for everyday users. 

However, despite their performance, VLA models face several challenges that limit their accessibility and widespread adoption. These models are often closed-source, computationally intensive (making them incompatible with consumer-grade GPUs), and require vast amounts of training data. As a result, their development and deployment are typically confined to organizations with substantial computational and financial resources. Additionally, VLA models are often tailored to specific robots, meaning that adapting them to new robots or tasks typically requires significant effort and retraining.

To address some of these challenges, the OpenVLA \cite{openvla} model has been proposed as an open-source alternative, offering greater compatibility. However, it still requires retraining for the addition of new robotic tasks or the adaptation to different robots.

In response to these limitations, we introduce SVLR (Scalable Visual Language Robotics), a scalable, training-free, and open-source framework designed to run on consumer-grade GPUs.
The SVLR framework leverages multiple AI models to process visual inputs and language instructions, enabling robots to execute complex tasks. SVLR integrates a Visual Language Model (VLM), a zero-shot image segmentation model, a Large Language Model (LLM), and a sentence similarity model. These models collaboratively extract information from the robot’s environment and interpret language commands to generate appropriate sequences of actions for task execution. Unlike traditional VLA models that directly generate actions, SVLR uses a set of pre-programmed tasks with adjustable parameters, such as object positions, to drive the robot’s actions. This modular approach allows for easy expansion, enabling the integration of new tasks and robots simply by adding text descriptions and programs. SVLR acts as a high-level controller that coordinates decision-making in robot task execution, allowing users to program tasks that can be either deep learning-based or model-based, depending on specific requirements. This flexibility makes it adaptable to a wide range of robotic platforms and use cases.

In this study, we programmed several low-level tasks for the UR10 robot, including ``pick and place,'' ``move to,'' ``open the gripper,'' and ``close the gripper.'' Since this paper primarily serves as a demonstration of the modularity and scalability of the SVLR framework rather than an exhaustive exploration of diverse robotic tasks, the focus was placed on the pick-and-place task, which is commonly used in VLA benchmarks.

This system offers several advantages. Its modular design enables seamless integration of advancements in AI technology, such as updates to VLMs, segmentation models, LLMs, or sentence similarity models, ensuring long-term adaptability. Additionally, the framework’s task set concept enables the straightforward addition of any robotic task requiring object positions, making it applicable across diverse robotic platforms. This paper details the design, implementation, and evaluation of SVLR, highlighting its potential to address the challenges faced by traditional VLA models.

\section{RELATED WORKS} 
\textbf{Closed-source VLA} Among the most notable VLA systems are the Robotics Transformer models, such as RT-2 \cite{rt2}, developed by Google. These systems have achieved significant advancements in converting natural language instructions into motion control commands for complex robotic tasks. RT-2 leverages large pre-trained VLMs like PaLI-X \cite{palix} (55 billion parameters) and PaLM-E \cite{palmee} (12 billion parameters) as backbones, harnessing their vision and language understanding capabilities acquired from web-scale data. However, this performance comes at the cost of resource-intensive models that require extensive computational infrastructure, making them inaccessible for widespread use. Moreover, these models are limited by their reliance on training data and often fail to generalize to motions outside their training set.

\textbf{Open-source VLA} To address the closed nature and fine-tuning challenges of proprietary VLA systems, OpenVLA \cite{openvla} was introduced as a 7 billion parameters open-source alternative. Trained on 970,000 robotic episodes from the Open X-Embodiment dataset \cite{rt2x}, OpenVLA offers broader compatibility and the ability to control multiple robot types. However, like other VLA systems, its scalability remains a concern. Training such models on diverse datasets demands significant computational and data resources, raising questions about their applicability to more complex and dynamic robotic tasks.

\textbf{Combining VLMs with Segmentation Models} Beyond traditional VLA, research has explored combining VLMs with segmentation models for object detection and vision tasks. For example, Vision-Instructed Segmentation and Evaluation (VISE) \cite{fewshotvqa} redefines few-shot image classification and segmentation as a form of Visual Question Answering (VQA). By integrating VLMs with tools like the Segment Anything Model \cite{SegmentA}, VISE achieves state-of-the-art results without extensive fine-tuning. While VLMs offer more general capabilities compared to object detection methods like YOLO \cite{yolo}, they output text rather than pixel coordinates. Consequently, segmentation pipelines like CLIPSeg \cite{CLIPSeg} are necessary to obtain coordinates.

\textbf{Large Language Model for robotic motion generation} LLMs have recently been applied to generate robotic motions from natural language instructions. For instance, \cite{alter3} presents an innovative approach where GPT-4's \cite{gpt4} language understanding capabilities are harnessed to generate humanoid robot motions. The authors developed a framework that translates natural language descriptions into structured motion representations using GPT-4. However, this approach faces challenges such as instability in predictions when minor textual perturbations occur, and it may struggle to generate consistent outputs for semantically similar inputs. To address these challenges and ensure reliable task convergence, we incorporates a sentence similarity model. This model evaluates the LLM's output and compares it against a predefined set of tasks and associated parameters, effectively constraining the LLM's responses to align with known robotic actions and ensuring consistency in task execution.
Similarly, \cite{langreward} uses LLMs to convert user instructions into reward-specifying code for robotic tasks. These rewards are then optimized by a motion controller using Model Predictive Control (MPC) with MuJoCo \cite{mujoco}. While this method bridges high-level instructions and low-level actions, it relies on the accuracy of LLM-generated rewards. Misinterpretations or errors in reward formulation may compromise task safety, underscoring the need for careful integration of LLMs in safety-critical applications. To address this, we introduces the concept of a set of pre-programmed tasks, where users need to program their own robotic tasks with explicit safety constraints if needed. This approach also allows for customization and adaptability to diverse robotic platforms without requiring extensive retraining.

\textbf{Addressing the challenges with SVLR} In this section, we discussed related works in language instruction-based robotic control, highlighting promising results alongside significant drawbacks, particularly related to scalability, unpredictable behavior, and the data-intensive nature of large generative AI. To address these challenges, the SVLR framework adopts an approach that leverages pretrained AI models for their ability to perform tasks or make predictions on unseen classes without requiring additional training. This capability enables these models to generalize to entirely new tasks or concepts that were not explicitly included in their training data. With the emergence of lightweight generative AI models (under 7 billion parameters), we designed a framework that is both adaptable to current advancements and future innovations, ensuring the architecture remains upgradable. Furthermore, SVLR is designed to utilize pre-programmed tasks, where the reliability and safety guarantees are contingent on how the tasks are designed. By employing a model-based approach to program these tasks, we can ensure safety guarantees. Moreover, To ensure that our framework reliably converges to a pre-programmed task, we incorporate a sentence similarity model to analyze the output of the LLM, as detailed in the next chapter.

\section{THE SCALABLE VISUAL LANGUAGE ROBOTICS FRAMEWORK}
\begin{figure}[H]
  \centering
  \def\svgwidth{0.425\textwidth} 
  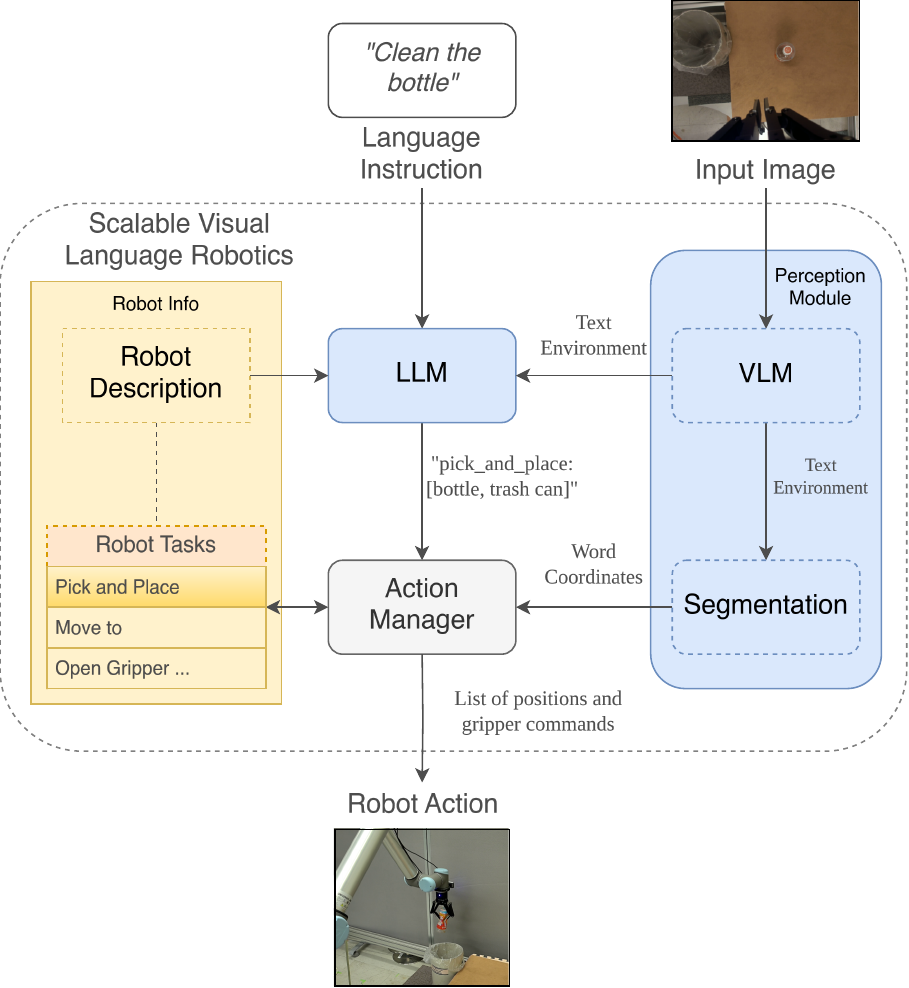 
  \caption{The Scalable Visual Language Robotics framework architecture}
  \label{fig:svlr}
\end{figure}

As illustrated in Fig.\ref{fig:svlr}, the SVLR framework requires two inputs: the user's language instruction and an image of the environment. These inputs are processed to generate a series of actions that the robot executes to follow the language instruction. 

The processing involves four primary components:

\begin{itemize}
    \item \textbf{Robot Info}: This component includes a text-based description of the robot's capabilities, provided to the LLM. It also contains the description and link to the robot tasks set, which is a collection of pre-programmed tasks. These tasks are selected and parametrized by the framework to execute the user's instruction.
    
    \item \textbf{Perception Module}: Represented by the VLM and segmentation model in Fig.\ref{fig:svlr}, this module identifies and retrieves objects in previously unseen environments.
    
    \item \textbf{Large Language Model}: The LLM determines the sequence of tasks required to respond to the language instruction. A prompt generator provides the LLM with the user's instruction, environmental information as perceived by the Perception Module, and details about the robot's capabilities as defined in the Robot Info module. 
    
    \item \textbf{Action Manager}: This component ensures alignment between the LLM's output and the perception module's data by employing a sentence similarity model. It also validates the tasks of the LLM's output against the robot tasks set. The Action Manager executes the finalized tasks with their parameters and sends commands to the robot controller.
\end{itemize}

Fig.\ref{fig:svlr} summarizes the relationships and data flow between these components. The following sections provide a detailed explanation of each component.

\subsection{Robot and tasks implementation}

Adding a robot to the SVLR framework requires a text description of the robot and a corresponding set of pre-programmed tasks.
This information is provided through a robot-specific JSON file that includes: the name and description of the robot, its initial position, the offset between the camera and the end effector, and the list of actions the robot can perform. Each task entry specifies the task name, description, associated program names, and parameters (if applicable).

The SVLR framework follows a client-server architecture to interact with the robot controller. It transmits the list of programmed tasks (output of each program) selected by SVLR to execute the user’s instruction.The system sends the controller a list of all the programmed tasks output, selected by SVLR to fulfill the language instruction. The robot controller is required to confirm task completion and ensure the robot returns to its initial position.
Further details and examples of programmed task outputs will be discussed in Section~\ref{sec:experiments}.

\subsection{Perception module}

The perception module comprises a VLM that generates descriptions of an image based on a given prompt, and a zero-shot segmentation model that segments the image according to VLM output. This module concludes by calculating the centroids of each detected object and converting their pixel coordinates to world coordinates.

With the recent emergence of open-source VLMs, we have opted for this type of model to benefit from general textual descriptions compared to traditional object detection solutions such as YOLO \cite{yolo}. As traditional VLMs generate only text (YOLO providing positions), the VLM's output is given to a zero-shot segmentation model to obtain the positions of detected objects in the image. This kind of architecture allows us to take advantage of the VLM's ability to generalize its detection to any type of environment. This approach also aligns with our objective of developing a modular system, as it offers users the flexibility to switch between different open-source models as needed.

The perception module takes an image of the environment as input. The VLM processes this image to generate a list describing each object present in the scene (refers as ``Text Environment'' in Fig.\ref{fig:svlr}). The list of detected objects is then passed to the segmentation model.

To guide the VLM in generating structured outputs, we use a simple but effective prompt: \textit{“List the objects, with only one object per line.”}. The VLM’s output may include additional details such as the color of the objects, which provide richer descriptions that can help the segmentation model to map the textual description of the object to pixel positions in the image.

\begin{figure}[H]
  \centering
  \includegraphics[width=0.425\textwidth]{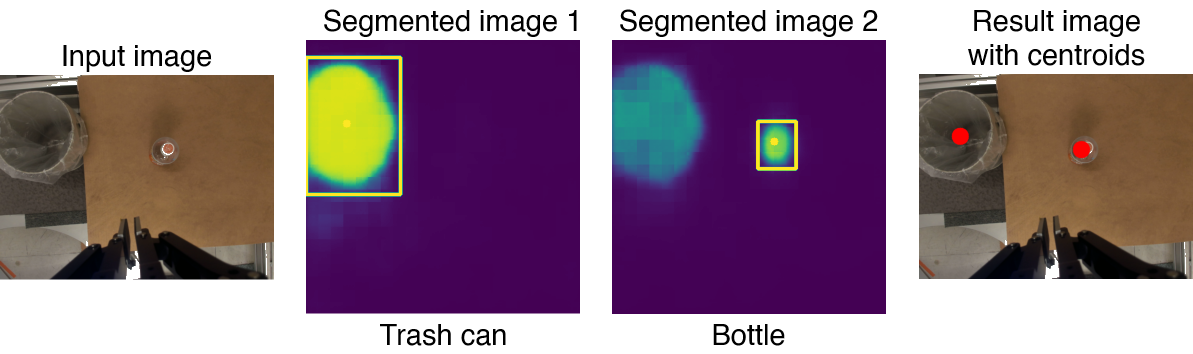}
  \caption{Example of the perception module's processing}
  \label{fig:segexample}
\end{figure}

In Fig.\ref{fig:segexample}, an example of the VLM output is shown, where the result is ``1. Trash can, 2. Bottle''. This output is converted into the list [''Trash can'', ``Bottle''], allowing the segmentation model to separately identify and locate the positions of each object in the image.

For each object's description in the list, the segmentation model generates a mask that outlines the shape corresponding to the given description within the image. Once the segmentation model produces a mask for each object, the centroid of the mask (geometric center) is computed. This is done by applying Gaussian blurring to the mask to reduce noise, followed by thresholding to create a binary image. Connected component analysis is performed to identify distinct regions within the mask, and the region with the highest median intensity is selected, which is often associated with the most significant or prominent feature in the image. The centroid is then calculated as the center of this selected region. Finally, the perception module translates the centroids' pixel coordinates into corresponding poses within the robot's coordinate system. This process involves using the depth value, which represents the distance between the camera and the objects, along with the camera's intrinsic parameters to convert the 2D pixel coordinates into normalized image coordinates. Then, the robot info module is used to apply the offset between the camera's position and the end effector. At the output of the perception module, we have the names of the objects in the environment and their world coordinates. 

The perception module combines VLMs and zero-shot segmentation to identify a wide range of objects. Unlike traditional object detection models trained on a fixed set of object classes, VLMs leverage their understanding of natural language and visual concepts to describe objects they were not explicitly trained on. This capability enables the perception module to generalize to virtually unlimited object categories, going beyond the constraints of traditional models. However, while the VLM excels at image description, it lacks the reasoning capabilities necessary for task planning. To address this, the LLM interprets the perception module's structured output and handles complex reasoning tasks, as discussed in the next section.

\subsection{Large Language Model and prompt generation}
\label{sec:llmandpg}

The aim of the LLM in the SVLR framework is to generate a series of actions from the robot's pre-programmed tasks set that effectively respond to the user's language instruction. To achieve this, we designed a prompt generator that exploits the following information:

\begin{itemize}
    \item \textbf{Prompt system template}: This component describes the expected behavior and output of the LLM. Since the prompt system may vary depending on the specific LLM used, we use a JSON file to define the prompt system for each LLM.
    
    \item \textbf{Text environment}: Given by the VLM, this provides the list of the object descriptions the robot can interact with, in text format.
    
    \item \textbf{Robot and task descriptions}: Derived from the robot info module, this includes information about the robot's type, its available actions, and the parameters required for those actions.
    
    \item \textbf{Language instruction}: This is the user's natural language command that the LLM must interpret.
\end{itemize}

Once the prompt is generated, the LLM processes it to produce a high-level textual description of the robot’s actions. For example, as shown in Fig.\ref{fig:svlr}, the instruction ``\textit{Clean the bottle}'' leads to the generation of the command: ``\textit{pick\_and\_place: [bottle, trash can]}''. While this example involves a single action, the framework supports more complex instructions resulting in multi-action outputs based on only one inference, as demonstrated in Section~\ref{sec:experiments}.

The benefits of using an LLM to control a robot extend beyond providing a natural language interface; it also enhances decision-making and adaptability, enabling the robot to adapt to various tasks and environments, allowing for adjustments based on specific circumstances  \cite{llmrobotics}.
For instance, in the example above, the user command does not explicitly mention the trash can or specify the pick-and-place action. The LLM deduces these based on the contextual information provided: the VLM identifies the trash can, and the robot info module specifies the robot’s pick-and-place capability.

While LLMs excel in generating flexible and adaptive high-level instructions, their reliability decreases for low-level control tasks such as trajectory generation or direct actuator commands, which are typically underrepresented in LLM training datasets \cite{langreward}. To mitigate this risk, the LLM's prompt must constrain LLM outputs to high-level task descriptions described in the following format: ``\textit{action\_name: [param\_1,... param\_n]}'' (e.g., ``\textit{pick\_and\_place: [bottle, trash can]}''), minimizing risks associated with misinterpretation. The LLM can generate as many high-level task descriptions as required to follow the language instruction, with each description separated by a line break. These high-level commands are later translated into low-level actions by the action manager, which executes the appropriate pre-programmed tasks and sends them to the robot controller.

\subsection{Action manager}
The action manager translates high-level commands from the LLM into executable robot control tasks. It parses the LLM's output line by line, extracting the action name and its associated parameters, which correspond to objects identified by the perception module. To ensure the correct parameter matching, the exact object names detected by the perception module are provided to the LLM as the text environment (see Section~\ref{sec:llmandpg}). However, discrepancies can still arise between the object names in the LLM's output and those detected by the perception module. For instance, while the perception module may detect a ``trash can'', the LLM might refer to it as ``trash\_can''. To resolve such cases, the action manager uses a sentence similarity model to match the parameters from the LLM's output to the closest corresponding objects detected by the perception module (e.g., perception output: ``trash can'', ``bottle'', ``cup''; parameter's name given by the LLM: ``trash\_can''; the similarity model maps ``trash\_can'' to ``trash can''). Once a match is established, the associated position coordinates from the perception module are assigned as the task parameters.

Similarly, the action manager uses the sentence similarity model to match the action names from the LLM output to the pre-defined actions described in the robot info module. After identifying the closest match, the action manager executes the task with the matched parameters and sends the result to the robot controller.

In the experiments, the output of pre-programmed tasks consists of end-effector positions of the robot is instructed to reach. The next section discusses these experiments.

\section{EXPERIMENTS}
\label{sec:experiments}
\subsection{Conducted Experiments}
To evaluate the performance of the SVLR framework, we conducted a series of experiments aimed at demonstrating the system's ability to execute language instructions. These experiments also tested the decision-making capabilities of the LLM and the perception module's effectiveness in detecting and handling previously unseen objects.

\paragraph{Hardware Setup}  
Experiments were conducted with a UR10 robot arm, a Robotiq 2F-140 gripper, and a standard webcam mounted on the gripper for image capture. The webcam lacked a depth sensor, so object depth had to be manually input. The SVLR framework ran on an NVIDIA RTX 2070 GPU (mobile) with 8GB VRAM, demonstrating the framework’s ability to operate on consumer-grade hardware. Note that runtime VRAM and latency depend on the AI models selected, not the SVLR framework itself. While small models were used to match the available hardware, users with larger models may need more powerful GPUs. However, with ongoing advancements in AI, the computational cost of running SVLR is expected to decrease as smaller models continue to perform similarly to larger ones.

\paragraph{Model Selection}  
The selection of AI models was guided by two key criteria: their relatively small size (fewer than 4 billion parameters) for efficient operation on consumer-grade GPUs and their strong performance on respective benchmarks at the time of testing.

\begin{itemize}
    \item \textbf{LLM}: microsoft/Phi-3-mini-4k-instruct, a 3.8B parameters model, quantized to 4 bits. \cite{Phi3}
    \item \textbf{VLM}: OpenGVLab/Mini-InternVL-Chat-2B-V1-5 \cite{mini-intern}, a 2B parameters model.
    \item \textbf{Segmentation}: CIDAS/clipseg-rd64-refined \cite{CLIPSeg}
    \item \textbf{Sentence similarity}: all-MiniLM-L6-v2 \cite{minilm}
\end{itemize}

To assess the scalability of the framework, we also tested cloud-based LLMs such as OpenAI's \textit{GPT-3.5} and \textit{GPT-4 }\cite{gpt4}. These models provided outputs comparable in accuracy to those generated \textit{by Phi-3} \cite{Phi3} in the SVLR context. However, due to the need for local deployment and resource accessibility, \textit{Phi-3} \cite{Phi3} was selected for the primary experiments, aligning with our goal of operating the system on consumer-grade hardware.

\paragraph{Experiment Setups}
The UR10 robot was programmed with four distinct actions that constitute the task set available to the SVLR framework: ``open the gripper,'' ``close the gripper,'' ``move to'' (where the robot moves to a position above a specified location), and ``pick and place''. Given the limitations of some actions, such as using the ``move to'' action that was only used when the user asked the robot to show an object, we decided to focus on the language instructions that use ``pick and place'' action for the remainder of this study. 

In our case, each action in the tasks set returns a list of end-effector positions along with the corresponding gripper values, indicating whether the gripper is open or closed. SVLR then sends this data directly to a ROS controller, which executes the list of actions on the robot without any additional processing.

The robot info module created for the UR10, along with its associated tasks set, is available in the GitHub repository of the project.

\subsection{Results}
\begin{figure}[H]
  \centering
  \includegraphics[width=0.425\textwidth]{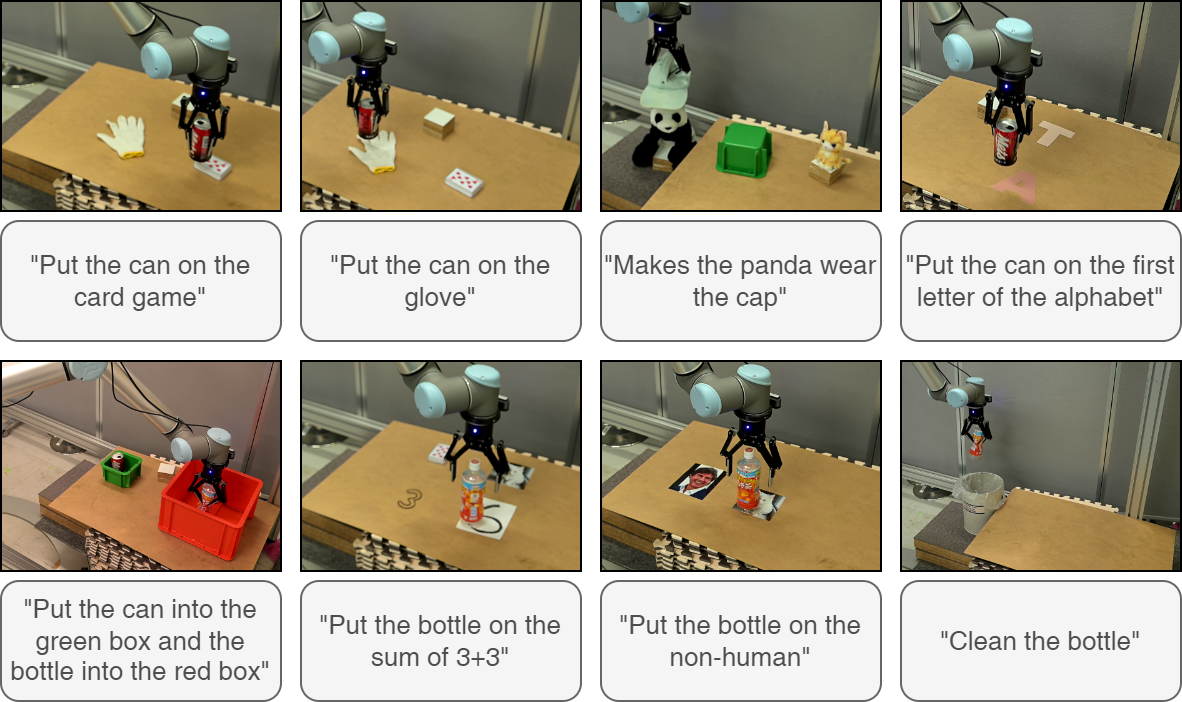}
  \caption{Different setups and language instructions used in the experiments}
  \label{fig:result}
\end{figure}
As shown in Fig.\ref{fig:result}, SVLR successfully performed a variety of tasks that required distinct reasoning and decision-making capabilities. The framework can handle complex and indirect requests. For instance, it can process queries such as identifying ``the first letter of the alphabet'', calculating ``the sum of 3+3''. Furthermore, it executed  higher-level commands such as ``clean,'' which necessitated a combination of reasoning and scene understanding. For instance, the system identified objects needing cleaning (the ``bottle''), determined the appropriate cleaning method (``pick and place''), and decided on disposal locations based on contextual information (the ``trash can''). SVLR also excels at performing multiple actions in response to a single request. For example, the instruction ``Put the can into the green box and the bottle into the red box'' results in two ``pick and place'' actions being generated.

Our perception module demonstrated robust object categorization, distinguishing between animals, inanimate objects, numbers, and letters. However, occasional inaccuracies occurred, such as the VLM misidentifying a panda as a rabbit. The framework can mitigates such errors using the LLM’s reasoning and the sentence similarity model to match the rabbit to the panda. This integration helps compensate for perception errors, ensuring the robot’s actions align with the intended outcomes.

\section{DISCUSSION AND LIMITATIONS}
We introduced SVLR, a scalable, training-free, lightweight, and modular alternative to traditional VLA. Despite using small, state-of-the-art generative AI models, significant results were achieved. However, the following sections outline potential improvements and future work.

\paragraph{Hardware and Software Improvements}  
Integrating a depth camera or multi-camera setup would improve environmental descriptions and enable more complex tasks, eliminating the need for arbitrary depth specification. Adding visual servoing could refine task execution by adjusting actions in real-time based on feedback.

Equipping the robot with a force-sensing gripper and accounting for object shape and orientation could improve pick-and-place efficiency. However, the focus of this work was to demonstrate the conceptual potential of a modular system, not to explore every task in detail.

\paragraph{Model Usage}  
AI models produce uncertain outputs. Although we used efficient small models at the time, advancements in the field suggest that future updates could reduce uncertainty and improve performance. As SVLR is built on a modular, multi-model approach, integrating newer models will enhance its robustness and efficiency over time.

\paragraph{Future Work}  
Conducting a benchmark by testing our system with various AI models, particularly lightweight ones, would offer valuable insights into performance variations and help assess the impact of different models on the overall efficiency of our system. This benchmarking should not only compare the success rate of our system with other VLA systems but also evaluate the performance of different combinations of VLMs, LLMs, segmentation, and sentence similarity models. Such comparisons will provide critical insights into the most effective configurations and guide further improvements to our architecture.

Furthermore, as SVLR is a high-level controller, we want to apply our framework to a diverse range of robots, and demonstrate its versatility across various applications and platforms. It would be particularly interesting to extend SVLR to mobile and humanoid robots, thereby showcasing its potential in configurations beyond robotic arms, which are currently over-represented in AI-based controllers such as VLA systems.

\section{CONCLUSION}

The Scalable Visual Language Robotics (SVLR) framework introduces a modular and training-free solution for robotic control, integrating both perception and reasoning pipelines. SVLR leverages a Vision Language Model (VLM) for general object recognition and a Segmentation model to convert text descriptions into world coordinates for robot control. The system’s reasoning pipeline, driven by a Large Language Model (LLM), uses the robot’s capabilities, as defined by the user through the robot info module and a custom task set, to generate a sequence of actions that fulfill the user's language requests. The integration of a sentence similarity model ensures that the LLM’s outputs match the predefined tasks and objects detected by the perception module. One of the key strengths of SVLR is its modular design, which allows for easy updates and integration of new AI models as they become available. By using pre-trained generative models, the framework enables users to customize their robot systems without the need for complex retraining. The user defines the tasks and robot capabilities textually, and SVLR’s modular architecture ensures smooth communication with robot controllers, such as the ROS client used in this study.

Our experiments with the UR10 robot arm have shown that SVLR can interpret and execute a variety of tasks, with encouraging performance. The framework's success on consumer-grade hardware underscores its accessibility and potential for widespread adoption. The results presented in this paper are a starting point, showing the potential of our approach. However, there are opportunities for further enhancement. While the development of compact generative models is still in its early stages, incorporating new models developed in the coming months could significantly enhance our outcomes. Future work should focus on benchmarking with a broader range of AI models to fine-tune performance and exploring additional robot types to extend the system’s versatility.

In summary, the SVLR framework not only highlights the potential of a modular, training-free AI-driven approach to robotic control but also paves the way for future developments that could expand its capabilities and applications across diverse robotic platforms and scenarios.

\addtolength{\textheight}{-12cm}   





\section*{ACKNOWLEDGMENT}
The authors thank the JRL team for insightful discussions and Thomas Duvinage for his assistance with the UR10 robot's controller implementation.



\bibliographystyle{ieeetr}
\bibliography{root}

\begin{thebibliography}{10}

\bibitem{rt2x}
E.~Collaboration, A.~O'Neill, A.~Rehman, {\em et~al.}, ``Open x-embodiment: Robotic learning datasets and rt-x models,'' 2024.

\bibitem{openvla}
M.~J. Kim, K.~Pertsch, {\em et~al.}, ``Openvla: An open-source vision-language-action model,'' 2024.

\bibitem{rt2}
A.~Brohan, N.~Brown, {\em et~al.}, ``Rt-2: Vision-language-action models transfer web knowledge to robotic control,'' 2023.

\bibitem{palix}
X.~Chen, J.~Djolonga, {\em et~al.}, ``Pali-x: On scaling up a multilingual vision and language model,'' 2023.

\bibitem{palmee}
D.~Driess, F.~Xia, {\em et~al.}, ``Palm-e: An embodied multimodal language model,'' 2023.

\bibitem{fewshotvqa}
T.~Meng, Y.~Tao, {\em et~al.}, ``Few-shot image classification and segmentation as visual question answering using vision-language models,'' 2024.

\bibitem{SegmentA}
A.~Kirillov, E.~Mintun, {\em et~al.}, ``Segment anything,'' 2023.

\bibitem{yolo}
J.~Redmon, S.~Divvala, {\em et~al.}, ``You only look once: Unified, real-time object detection,'' 2016.

\bibitem{CLIPSeg}
T.~Lüddecke and A.~S. Ecker, ``Image segmentation using text and image prompts,'' 2022.

\bibitem{alter3}
T.~Yoshida, A.~Masumori, {\em et~al.}, ``From text to motion: Grounding gpt-4 in a humanoid robot "alter3",'' 2023.

\bibitem{gpt4}
OpenAI, J.~Achiam, S.~Adler, S.~Agarwal, {\em et~al.}, ``Gpt-4 technical report,'' 2024.

\bibitem{langreward}
W.~Yu, N.~Gileadi, {\em et~al.}, ``Language to rewards for robotic skill synthesis,'' 2023.

\bibitem{mujoco}
T.~Howell, N.~Gileadi, {\em et~al.}, ``Predictive sampling: Real-time behaviour synthesis with mujoco,'' 2022.

\bibitem{llmrobotics}
F.~Zeng, W.~Gan, {\em et~al.}, ``Large language models for robotics: A survey,'' 2023.

\bibitem{Phi3}
M.~{Abdin}, J.~{Aneja}, {\em et~al.}, ``{Phi-3 Technical Report: A Highly Capable Language Model Locally on Your Phone},'' {\em arXiv e-prints}, p.~arXiv:2404.14219, Apr. 2024.

\bibitem{mini-intern}
Z.~Chen, J.~Wu, {\em et~al.}, ``Internvl: Scaling up vision foundation models and aligning for generic visual-linguistic tasks,'' 2024.

\bibitem{minilm}
HuggingFace, N.~Reimers, J.~Gante, {\em et~al.}, ``all-minilm-l6-v2: https://huggingface.co/sentence-transformers/all-minilm-l6-v2,'' 2021.

\end{thebibliography}
\end{document}